\documentclass[runningheads]{llncs}

\usepackage{eccv}

\usepackage{eccvabbrv}

\usepackage{graphicx}
\usepackage{booktabs}

\usepackage[accsupp]{axessibility}  

\usepackage{subcaption} 
\usepackage{amsmath}
\usepackage{amssymb}
\usepackage{multirow}
\usepackage{xspace}
\usepackage{xcolor}

\usepackage{hyperref}

\usepackage{orcidlink}

\begin{document}

\title{Environmental Change Detection for \\ Real-World Change Analysis} 

\titlerunning{Environmental Change Detection}

\makeatletter
\newcommand{\samecorr}{\textsuperscript{\@fnsymbol{1}}}
\makeatother

\author{
Kyusik Cho\inst{1}\orcidlink{0009-0008-7205-1700} \and
Suhan Woo\inst{1}\orcidlink{0009-0008-1947-750X} \and
Hongje Seong\inst{2}\samecorr\orcidlink{0000-0001-7221-409X} \and
Euntai Kim\inst{1,3}\thanks{Corresponding authors.}\orcidlink{0000-0002-0975-8390}
}

\authorrunning{K. Cho et al.}

\institute{\textsuperscript{\rm 1} Yonsei University \quad \textsuperscript{\rm 2} University of Seoul \\
\textsuperscript{\rm 3} Korea Institute of Science and Technology\\
\email{\{ks.cho, wsh112, etkim\}@yonsei.ac.kr} \quad \email{hjseong@uos.ac.kr}
\url{https://kyusik-cho.github.io/ECD}
}

\maketitle

\begin{abstract}
Scene Change Detection (SCD) evaluates changes using predefined query-reference (\ie, present-past) image pairs.
However, this formulation overlooks a critical dependency: the corresponding query-reference pair is assumed to be prepared in advance.
In real-world applications, such as mobile robots, future query views cannot be known in advance, and thus their corresponding reference images cannot be predefined.
To remove this dependency and push change detection toward more practical applications, we introduce Environmental Change Detection (ECD).
A key aspect of ECD is to avoid unrealistically predefined and aligned query-reference pairs and instead retrieve environmental cues from an uncurated image database of reference scenes.
To tackle this new challenging task, we additionally introduce an initial solution that enables change detection under unknown and imperfect query-reference conditions.
The main idea of our solution is to retrieve multiple reference candidates and aggregate semantically rich representations for change detection.
We further construct ECD benchmark sets by reformulating three standard change detection datasets.
Extensive experimental results demonstrate the efficacy of our solution in both ECD and SCD.
\end{abstract}

\begin{figure}[!t]
\centering
\includegraphics[width=1.0\linewidth]{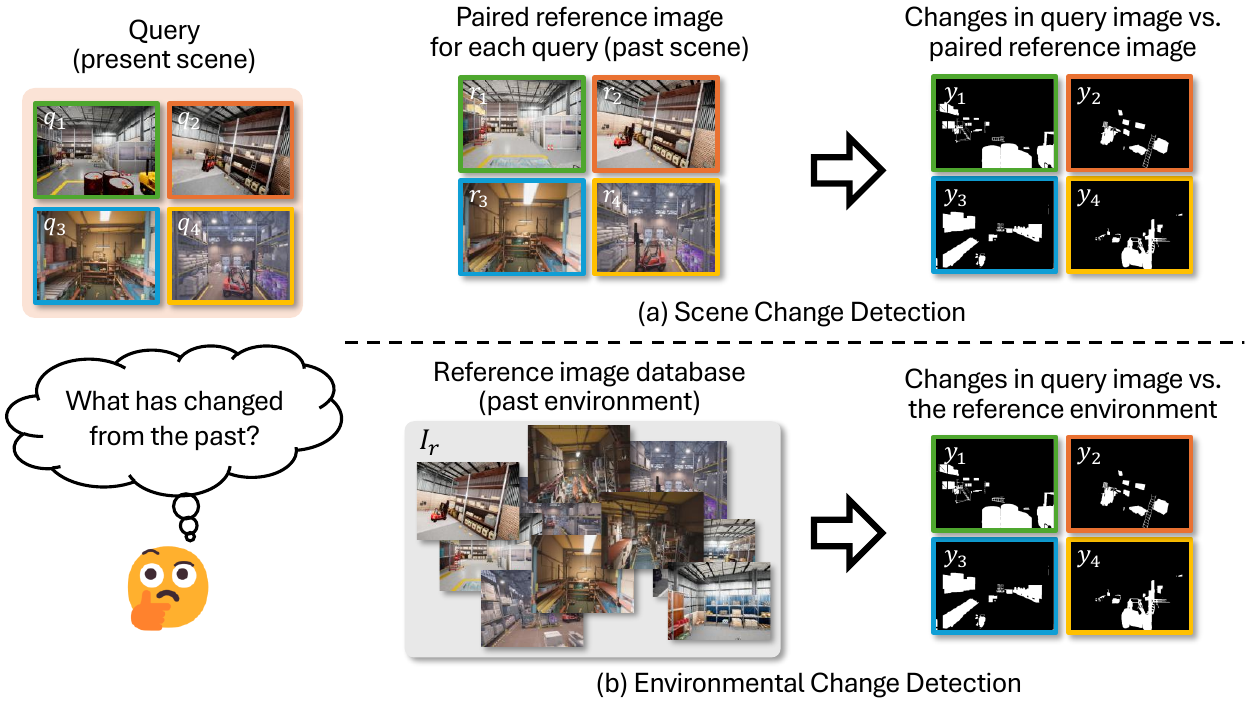}
\caption{
Comparison between the problem settings of SCD and ECD.
In the conventional SCD setting, each query image is provided with its corresponding reference image as a predefined pair. In contrast, ECD considers a more practical scenario where query-reference pairs are unknown.
ECD just provides a single query image, where the database of reference images is accessible. Furthermore, ECD does not guarantee the presence of a perfectly matched pair within the database.
}
\label{fig:task}
\end{figure}

\section{Introduction}
\label{sec:intro}
Change detection in images is a fundamental problem in computer vision. Traditionally, this problem has been addressed through Scene Change Detection (SCD), which aims to identify differences between a reference image and a query image captured at distinct times.
SCD has attracted increasing attention due to its potential in a wide range of practical applications, such as
urban scene analysis~\cite{alcantarilla2018street, sakurada2020weakly, alpherts2025emplace}, disaster damage assessment~\cite{jst2015change, ahn2025generalizable}, anomaly detection~\cite{li2024umad}, and warehouse management~\cite{park2021changesim,park2022dual}.

Despite its significance, SCD has limited applicability in real-world scenarios. 
This limitation arises primarily from two idealized assumptions: (1) an appropriate reference image for every query is predefined before the query is presented, and (2) each query-reference pair is captured from the same viewpoint, ensuring spatial alignment.
Several recent studies, such as SimSac~\cite{park2022dual} and RSCD~\cite{lin2024robust}, have attempted to relax the second assumption by handling misalignments through optical flow or cross-attention.
However, these approaches still rely on the predefined pairing assumption, which imposes an inherent limitation: unobservable regions caused by viewpoint mismatch impose an upper bound on achievable performance.
Consequently, such extensions toward mismatch handling do not fully resolve the practicality issues of the conventional SCD formulation.

To address these limitations, we introduce Environmental Change Detection (ECD).
ECD is inspired by how changes are detected in real-world scenarios.
To detect changes from the past to the present, an appropriate past image does not emerge simultaneously with the query; rather, only a database of past images exists.
In such cases, humans first search for relevant images from the past image database.
In reality, perfectly aligned past images are exceedingly rare, and the available images often have only a limited field-of-view overlap.
Under these challenging conditions, humans fill in missing information from multiple observations rather than abandoning inference for unobservable regions.

ECD is designed to reflect these common practices.
In ECD, the reference information is indirectly provided by a large-scale reference database rather than predefined pairs, and a perfectly aligned image may not exist within it.
This formulation simultaneously relaxes both assumptions inherent in SCD, thereby making the task more challenging while improving its relevance to real-world applications.
The comparison between SCD and ECD is illustrated in \cref{fig:task}.

Conceptually, ECD can be viewed as a large-scale multi-reference change detection problem with many query-irrelevant references acting as distractors.
To address this setting, we propose a framework that identifies and integrates query-relevant reference evidence for change detection.
Directly processing the entire reference database is computationally prohibitive and exposes the detector to numerous irrelevant references.
Instead, our method first uses VPR to construct a compact, query-specific subset of informative reference candidates.
Then, we introduce the Environment Mosaic Composer (EMC), which reconstructs a query-oriented reference environment representation by selectively composing the most relevant feature patches for each region of the query in a mosaic fashion.
After applying EMC across multiple scales, we aggregate the resulting representations to further enhance robustness.
A change segmentation head is subsequently applied to produce the final change map.

Furthermore, we reformulate three standard change detection datasets and construct ECD benchmark sets.
Experimental results highlight the effectiveness of the proposed approach and provide a step toward practical applications of change detection.
Our contributions are summarized as follows:

\begin{itemize}
    \item We identify practical limitations of the existing change detection approaches and introduce a new challenging task named ECD.
    \item We propose the first solution to ECD, which predicts changes by retrieving and integrating multiple reference candidates, inspired by human reasoning.
    \item We establish standard experimental settings for ECD, providing benchmark datasets and an initial baseline to guide further studies.
\end{itemize}

\section{Related Work}
\paragraph{Scene change detection.}
Recent research on scene change detection (SCD) has leveraged deep learning to develop a variety of approaches~\cite{ramkumar2021self,sachdeva2023change,lee2024semi}.
DR-TANet~\cite{chen2021dr} introduced the temporal attention module into the encoder-decoder architecture.
C-3PO~\cite{wang2023reduce} proposes a model that utilizes temporal features to distinguish three types of changes.
EMPLACE~\cite{alpherts2025emplace} performs self-supervised learning using a time-interval-based triplet loss. 
Sachdeva and Zisserman~\cite{sachdeva2023change} and Lee and Kim~\cite{lee2024semi} utilized synthetic change data. 
ZSSCD~\cite{cho2025zero} leverages a foundation tracking model to perform change detection without training. 
Meanwhile, several methods are designed to be robust against misalignment between the reference and query images. 
SimSac~\cite{park2022dual} incorporates optical flow estimation and a warping module to account for misalignment.
RSCD~\cite{lin2024robust} introduces a cross-attention-based change detection head, enabling comparison even when object positions differ. 
However, these methods still rely on the predefined query-reference pairs.
When misalignment is introduced under such predefined pairing, unobservable regions inevitably arise, making the formulation impractical.
Hence, it is essential to resolve both problems simultaneously.
To jointly overcome these limitations, we establish a novel setting that removes the assumptions of predefined pairing and spatial alignment.

\begin{figure*}[!t]
\centering
\includegraphics[width=1.0\linewidth]{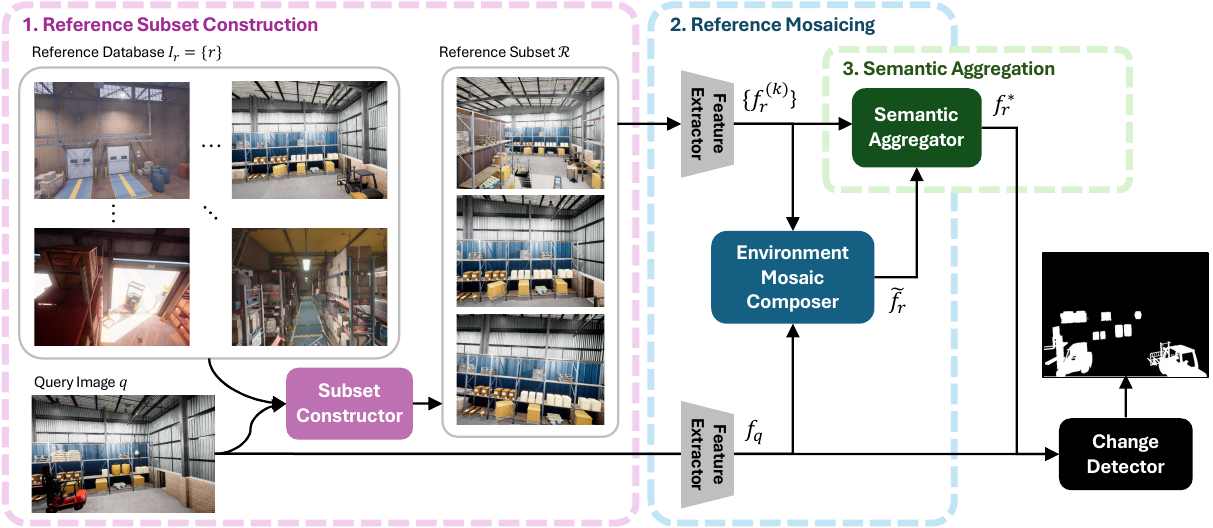}
\caption{Overview. 
Our approach first constructs a query-specific subset from the reference database.
Then, EMC generates a mosaic reference environment representation that matches the perspective of the query image. 
Based on the mosaic representation, the semantic aggregator integrates the semantic information across the subset. 
Finally, both the reference features and the query features are fed into the change detector to produce the final prediction.
}
\label{fig:overview}
\end{figure*}

\paragraph{Visual place recognition.} 
Visual place recognition (VPR) is the task of estimating the location of a query image by retrieving visually similar images from a pre-collected, geo-tagged database. 
With the rapid progress of deep learning, numerous effective methods have been proposed to tackle this challenge. 
NetVLAD~\cite{arandjelovic2016netvlad} is one of the first works to successfully apply deep learning to VPR, outperforming traditional SIFT-based methods~\cite{lowe1999object} and becoming a common baseline in later studies~\cite{qiu2024emvp,lu2024supervlad,zhu2023r2former,wang2022transvpr,lu2024cricavpr}. 
DINOv2 SALAD~\cite{izquierdo2024optimal} reformulates local feature aggregation in VPR as an optimal transport problem using the Sinkhorn algorithm, and employs DINOv2~\cite{oquab2023dinov2} as the backbone, achieving competitive performance with reduced training cost.
BoQ~\cite{ali2024boq} trains learnable global queries that aggregate local features via cross-attention to enable accurate and fast retrieval.
Meanwhile, several SCD studies have been associated with VPR. 
For example, SimSac~\cite{park2022dual} utilized VPR to prepare imperfect match pairs for SCD.
ZeroSCD~\cite{kannan2024zeroscd} leveraged a VPR backbone for feature extraction in SCD, inspired by the observation that VPR models are inherently robust to variations in style and content.
Unlike previous approaches that use VPR to prepare image pairs~\cite{park2022dual} or extract useful representations~\cite{kannan2024zeroscd}, we use VPR as the first stage of our solution to construct a query-specific reference subset from a large-scale reference database.

\section{Environmental Change Detection}
\subsection{Background: Scene Change Detection (SCD)}
Conventional SCD datasets~\cite{alcantarilla2018street, sakurada2020weakly, park2021changesim} consist of \((r, q, y)\) triplets, 
where \(r\) and \(q\) are the pre-change (\ie, reference) and post-change images (\ie, query) taken at time \(t_0\) and \(t_1\), respectively, and \(y\) denotes the corresponding ground-truth change map. 
The objective is to predict the pixel-wise change mask \(\hat{y}\) for the given pair \((r, q)\), formulated as
     $ \hat{y} = F (r, q), $
where $F$ denotes the SCD system.
This formulation is based on two idealized assumptions:
(1) every query image \(q\) is paired with a corresponding reference image \(r\), and
(2) the image pair \((r,q)\) is spatially aligned. 
Although these assumptions substantially simplify the change detection problem, they limit its broader applicability in practical real-world scenarios.

\subsection{Problem Formulation}
To increase its applicability in practical environments, we introduce Environmental Change Detection (ECD).
\cref{fig:task} contrasts SCD with ECD and highlights their key differences.
ECD relaxes the two unrealistic assumptions of SCD as follows:
(1)~ECD replaces the predefined paired input \((r, q)\) with a query-database input \((\mathcal{I}_r,q)\), where \(q\) denotes the query image and \(\mathcal{I}_r = \{r^{(i)}\}_{i=1}^{N_r}\) denotes a large-scale reference image database. 
Importantly, \(\mathcal{I}_r\) is not constructed specifically for each query or sequence. 
Instead, the same reference database is shared across all queries and sequences and may therefore contain images captured in environments entirely different from that of the current query.
The system must explore \(\mathcal{I}_r\) and collect useful reference information relevant to the given \(q\), reflecting practical usage scenarios.
(2)~A reference image \(r\) that is perfectly aligned with \(q\) is not guaranteed to exist in \(\mathcal{I}_r\).
To emulate this condition, we define \(\mathcal{I}_r\) as a uniformly sampled subset of the entire collection of reference images. 
The goal of ECD is to predict a pixel-wise change mask~\(\hat{y}\) for a given \(q\) by leveraging the information within~\(\mathcal{I}_r\), formulated as
      $\hat{y} = G (\mathcal{I}_r , q ),$
where $G$ denotes the ECD system.

\section{Method}
The overall architecture of our ECD framework is illustrated in \cref{fig:overview}.
To explore \(\mathcal{I}_r\), we first employ a VPR-based retrieval scheme to construct a compact subset of relevant reference images.
Next, we apply the Environment Mosaic Composer (EMC) to the constructed subset.
EMC reconstructs a mosaic representation of the environment by selectively composing feature patches from multiple references to match the spatial layout of the query \(q\).
Subsequently, semantic features from the subset are aggregated based on this mosaic representation to further enrich the reference understanding.
Finally, change detection is performed using the enriched reference representation and the query.
    
\subsection{Reference Subset Construction}
Instead of directly utilizing the entire reference database $\mathcal{I}_r$, we adopt a selective strategy that focuses on the most relevant subset for each query $q$. 
Our subset constructor consists of two components: a VPR module and a subset size parameter $K$.
The VPR module computes place-level similarity between the query image 
$q$ and all reference images in $\mathcal{I}_r$, and the top-$K$ images form the reference subset $\mathcal{R}$:
\begin{equation}
\mathcal{R}=\{r^{(1)},r^{(2)}, \dots ,r^{(K)} \} = \operatorname{TopK}_{\text{VPR}}(q, \mathcal{I}_r).
\end{equation}
This subset serves as a compact and relevant representation of $\mathcal{I}_r$ for estimating changes with respect to $q$. 

According to the ECD formulation, however, each retrieved image 
$r^{(k)} \in \mathcal{R}$
is highly likely to be captured from a viewpoint different from that of $q$.
Consequently, simply relying on a single top-ranked reference often results in suboptimal performance due to viewpoint misalignment and limited scene coverage.
While conventional SCD approaches attempt to alleviate the mismatch between paired images, they inherently overlook the fundamental limitation that a single reference image can only represent a narrow portion of the environment.
To overcome this, we introduce a new paradigm that composes the regions most relevant to the query from multiple viewpoints, thereby enabling a query-oriented understanding of the scene beyond the constraints of any single reference.

\subsection{Environment Mosaic Composition}
We introduce a new module, termed Environment Mosaic Composer (EMC), which reconstructs a query-oriented reference environment representation inspired by the way humans mentally reconstruct spatial layouts from fragmented observations.
When no single image fully captures the query viewpoint, humans imagine the missing parts of a scene.
They intuitively assemble the most plausible regions from multiple viewpoints to complete the query view, forming a coherent mental picture.
Analogously, EMC divides the query feature map into local patches and selects the most relevant reference feature patch for each patch.
Without performing any explicit pixel-level reconstruction, this compositional process yields a mosaic environment representation \( \tilde{f_r} \).
The resulting representation provides sufficient spatial correspondence cues for reliable change detection.

To realize this composition process, all query and reference images are first encoded using a frozen DINO backbone~\cite{oquab2023dinov2} to produce feature maps \( f_q \) and \( \{f_r^{(k)}\}_{k=1}^{K} \). 
These features are then \(\ell_2\)-normalized along the feature dimension.
Next, the query feature \( f_q \) is divided into an \( n \times n \) grid, forming the set of grid locations \( \mathcal{P}_n \). 
For each grid cell \( p \in \mathcal{P}_n \), we obtain a local patch \( f_q[p] \in \mathbb{R}^{d \times h \times w} \), where $d$, $h$, and $w$ denote channel, height, and width dimensions, respectively. Each query patch is compared with all possible patches from the reference features: 
\begin{align}    
S_k^p &=  f_r^{(k)} \circledast f_q[p],   \\
(k^*, z^*) &= \arg\max_{k, z} S_k^p[z] \; ,
\end{align}
where \( \circledast \) denotes 2D convolution over the spatial dimensions, computing a sliding-window sum of local cosine similarities.
The index $z$ ranges over all valid stride-$1$ locations in $f_r^{(k)}$ at which a reference patch with the same spatial dimensions as $f_q[p]$ can be extracted.
The patch with the highest similarity $f_r^{(k^*)}[z^*]$ is then copied from the reference map and placed into the mosaic, composing a query-oriented reference environment representation:
\begin{equation}    
\tilde{f}_{r,n}[p] = f_r^{(k^*)}[z^*] \; .
\end{equation}

\begin{figure}[!t]
\centering
\includegraphics[width= 1.0\linewidth]{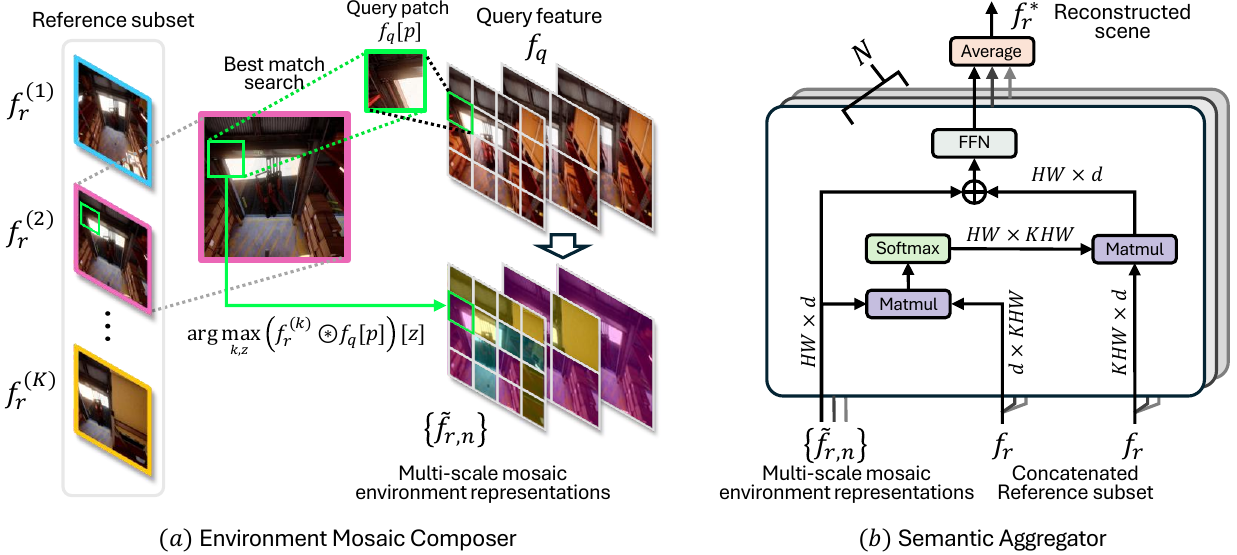}
\caption{Illustration of environment mosaic composer and semantic aggregator. 
(a)
The input query feature $f_q$ is divided into an $n \times n$ grid.
EMC searches over all reference features $\{f_r^{(k)}\}_{k=1}^K$ and identifies the best-matching patch through a convolution-based similarity computation.
By compositing the best matches across all grid cells, EMC constructs a mosaic environment representation $\tilde{f}_{r,n}$ that reflects the viewpoint of the query. This process is performed hierarchically with multiple $N$ scales to capture both coarse and fine-grained correspondences.
(b)
Features from the reference subset are integrated via cross-attention, using the mosaic environment representation $\tilde{f}_{r,n}$ as the query. Semantic aggregation is performed at each grid resolution, and the resulting features are averaged to produce the final reconstructed scene $f_r^*$.
For clarity, the illustration of the multi-head attention (MHA) is omitted.
}
\label{fig:detailed}
\end{figure}

This process is repeated for all \( p \in \mathcal{P}_n \) and across multiple grid resolutions $n$.
Consequently, multi-scale mosaic environment representations \( \{\tilde{f}_{r,n}\} \), which reflect the spatial layout of \( f_q \), are obtained. 
An illustration of EMC is provided in \cref{fig:detailed}a.

\subsection{Semantic Aggregation and Change Segmentation}
Each mosaic environment representation \( \tilde{f}_{r,n} \) provides strong spatial alignment with the query feature \( f_q \), but the compositional process may introduce spatial discontinuities at patch boundaries, which can lead to minor semantic inconsistencies across adjacent regions. 
To complement such inconsistencies and enrich the contextual understanding of the reconstructed scene, we incorporate a semantic aggregator that refines information from multiple references through cross-attention.
The strong spatial correspondence of \( \tilde{f}_{r,n} \) makes it particularly suitable to serve as the query in the cross-attention mechanism.
Meanwhile, the reference features \( \{ f_r^{(k)} \}_{k=1}^K \) offer complementary contextual cues and are thus employed as the keys and values.
We implement this process using a multi-head attention~\cite{vaswani2017attention}, followed by dropout, a residual connection, and a two-layer feed-forward network (FFN) with ReLU activation:
\begin{equation}
f_{r,n}^* = \text{FFN}(\text{Dropout}(\text{MHA}(\tilde{f}_{r,n}, f_r, f_r)) + \tilde{f}_{r,n}) \; ,
\end{equation}
where \( f_r = \text{concat}(f_r^{(1)}, \dots, f_r^{(K)}) \).

The resulting representation \( f_{r,n}^* \) is semantically enriched and spatially aligned with \( f_q \) at grid resolution $n$.
Finally, we generate the reconstructed scene \( f_r^* \) by averaging the representations \( f_{r,n}^* \).
An illustration of the semantic aggregator is provided in \cref{fig:detailed}b.

To predict the change, we feed \( f_r^* \) and \( f_q \) into a change detection module.
Following RSCD~\cite{lin2024robust}, we adopt a change detection module consisting of a shallow cross-attention block followed by convolutional layers for pixel-level change prediction.
The architectural details are provided in the supplementary material.

\section{Experiments}
\subsection{Datasets}
We select three SCD datasets, VL-CMU-CD~\cite{alcantarilla2018street}, PSCD~\cite{sakurada2020weakly}, and ChangeSim~\cite{park2021changesim}, and restructure them for ECD.
The key modification in our restructuring is the removal of the conventional query-reference pairing assumption.
Specifically, the dataset is no longer composed of paired tuples \((r, q)\) but instead two independent sets: a set of reference images \(\mathcal{I}_r\) and a set of query images \(\mathcal{I}_q\).
To further eliminate the unrealistic assumption that a spatially aligned reference image \(r\) always exists in \(\mathcal{I}_r\) for every \(q \in \mathcal{I}_q\), we define \(\mathcal{I}_r\) as a subset sampled from the entire collection of pre-change (reference) images of the original dataset.
The sampling is performed by applying a database stride \(s\), resulting in \(\mathcal{I}_r\) being downsampled to approximately $1/s$ of its original size.
This increases both the practicality and difficulty of the benchmark, approximating realistic scenarios in which past environments are sparsely captured with camera-mounted vehicles.
We generally write the restructured dataset as VL-CMU-CD$^s$, PSCD$^s$, and ChangeSim$^s$, and when a specific value is used, we indicate it explicitly (\eg, ChangeSim$^5$ for $s=5$). In our experiments, due to the maximum sequence lengths of VL-CMU-CD and PSCD (20 and 15, respectively), we consider $s$ up to 10; however, if experiments are conducted on datasets with longer sequences, this setting can be naturally extended beyond 10.

\cref{tab:ecd_statistics} summarizes the statistics of the test sets in the restructured ECD datasets under different database strides.
To characterize the geometric difficulty induced by reference sparsification, we report two overlap measures: reference-reference ($r$-$r$) overlap and query-reference ($q$-$r$) overlap. We define overlap as the percentage of one image that is geometrically visible in another image. The $r$-$r$ overlap measures the average geometric overlap between reference images in the downsampled reference database, while the $q$-$r$ overlap measures the overlap between each query and its top-1 reference. For PSCD, the overlap is computed directly from the panoramic crop geometry. For VL-CMU-CD and ChangeSim, it is estimated using a RANSAC-based homography~\cite{fischler1981random} fitted to LoFTR matches~\cite{sun2021loftr}. As the stride increases, both the reference database size and the overlap generally decrease, indicating that a single reference image provides increasingly limited geometric coverage.

VL-CMU-CD$^s$ is a restructured dataset derived from VL-CMU-CD~\cite{alcantarilla2018street}, which captures changes in urban street scenes. The dataset primarily represents changes such as the disappearance of buildings, roads, and vehicles caused by long temporal intervals. 
Following the previous work~\cite{lin2024robust}, we resize the images to $504\times504$, and split the training dataset into 3,324 queries for training, 408 for validation.

PSCD$^s$ is a restructured dataset of PSCD~\cite{sakurada2020weakly}, which consists of panoramic urban scene images. Following the previous work, we crop each panoramic image into 15 sub-images with a size of $224\times224$ and consider them as a sequence. PSCD$^s$ is used only for testing purposes, and evaluations are performed using the best model obtained from training and validation on VL-CMU-CD$^s$.

ChangeSim$^s$ is a restructured dataset derived from ChangeSim~\cite{park2021changesim}, which simulates an industrial warehouse environment. The dataset includes changes such as the appearance, disappearance, and rotation of various objects.
It contains 13,225 coarsely aligned image pairs for training and 8,212 for testing.
We split the original training set into 8,754 training and 4,471 validation samples.
Each image is resized to $224\times224$. 
In ChangeSim, each reference image is sampled from long sequences. As a result, multiple query images may share the same reference image, and some reference images may not be used at all. Therefore, rather than collecting reference images from the paired structure, we directly utilize the full $t_0$ sequences as a pre-change database before applying \(s\).

\begin{table}[!t]
\centering
\setlength{\tabcolsep}{2pt} 
\caption{Statistics of the restructured ECD datasets under different database strides.}
\label{tab:ecd_statistics}
\resizebox{1.0\columnwidth}{!}{
\begin{tabular}{@{}c|ccc|cc|c |ccc|cc|c |ccc|cc|c@{}}
\toprule
  & \multicolumn{6}{c|}{VL-CMU-CD$^s$}   & \multicolumn{6}{c|}{PSCD$^s$}  & \multicolumn{6}{c}{ChangeSim$^s$}   \\
  & \multicolumn{3}{c|}{Database size}  & \multicolumn{2}{c|}{Overlap (\%)} & \raisebox{-2pt}{\multirow{2}{*}{\shortstack{~avg.~\\[-2pt]seq. len.}}}
  & \multicolumn{3}{c|}{Database size}  & \multicolumn{2}{c|}{Overlap(\%)} & \raisebox{-2pt}{\multirow{2}{*}{\shortstack{~avg.~\\[-2pt]seq. len.}}}
  & \multicolumn{3}{c|}{Database size}  & \multicolumn{2}{c|}{Overlap(\%)} & \raisebox{-2pt}{\multirow{2}{*}{\shortstack{~avg.~\\[-2pt]seq. len.}}}   \\
$s$  
& ~\( |\mathcal{I}_q|\)~ &  ~\(|\mathcal{I}_r^{s}|\)~ &  \(|\mathcal{I}_r^{s}| / |\mathcal{I}_q|\)  & $r$-$r$ & $q$-$r$ &   
& ~\( |\mathcal{I}_q|\)~ &  ~\(|\mathcal{I}_r^{s}|\)~ &  \(|\mathcal{I}_r^{s}| / |\mathcal{I}_q|\)  & $r$-$r$ & $q$-$r$ &
& ~\( |\mathcal{I}_q|\)~ &  ~\(|\mathcal{I}_r^{s}|\)~ &  \(|\mathcal{I}_r^{s}| / |\mathcal{I}_q|\)  & $r$-$r$ & $q$-$r$ &
\\ 
\midrule
1     &   429 & 429 & 1.00    &    97.28 & 96.90    &   7.94     &     1155 & 1155 & 1.00    &  74.55 & 100.00 & 15           &     8212 & 5910 & 0.72   &    98.51  & 96.11  &  738.75         \\ 
3     &   429 & 162 & 0.38    &    96.45 & 95.99    &   3.00     &     1155 & 385  & 0.33    &  23.66 & 81.34  & 5            &     8212 & 1973 & 0.24   &    95.72  & 95.13  &  246.63         \\ 
5     &   429 & 106 & 0.25    &    94.65 & 94.54    &   1.96     &     1155 & 231  & 0.20    &  ~0.00 & 62.80  & 3            &     8212 & 1185 & 0.14   &    93.37  & 94.54  &  148.13        \\ 
10    &   429 & 66  & 0.15    &    92.08 & 93.86    &   1.22     &     1155 & 154  & 0.13    &  ~0.00 & 42.80  & 2            &     8212 & 596  & 0.07   &    88.47  & 94.44  &   74.50        \\ 
\bottomrule
\end{tabular}
}
\end{table}

\subsection{Experimental Setup}
\paragraph{Training details.} 
Following previous work~\cite{lin2024robust}, we use the Adam optimizer~\cite{kingma2015adam} with a learning rate of \(1 \times 10^{-4}\) and apply a cosine learning rate decay schedule~\cite{loshchilov2016sgdr}.
The model is trained for 100 epochs with a batch size of 4, including 10 warm-up epochs, and using the weighted cross-entropy loss. 
The VPR model~\cite{ali2024boq} and DINO feature extractor~\cite{oquab2023dinov2} are kept frozen throughout the pipeline. 
We set the reference subset size to $K=3$ by default, and define the EMC multi-scale grid resolution as $n \in \{1,2,4\}$.

\paragraph{Evaluation metrics.} 
Following previous work, we employ the F1-score as the evaluation metric for all datasets. 

\begin{table}[!t]
\centering
\setlength{\tabcolsep}{4pt}  
\caption{Experimental results on the ECD benchmark. Our method consistently outperforms the state-of-the-art change detector across all datasets and database stride $s$ settings. 
For a fair comparison, we feed reference images retrieved from the VPR model~\cite{ali2024boq} to RSCD.
}
\label{tab:main}
\begin{tabular}{cccccc}
\toprule
\multirow{2}{*}{~~$s$~~} & \multirow{2}{*}{~~~Method~~~} & \multicolumn{3}{c}{Dataset}      & \multirow{2}{*}{~Average~}    \\
                      & & VL-CMU-CD$^s$ & ~PSCD$^s$~ & ChangeSim$^s$  \\                
\midrule
\multirow{2}{*}{1} & RSCD~\cite{lin2024robust}   & 0.619            & 0.300           &   0.369            & 0.429    \\
 & Ours                        & \textbf{0.704}   &  \textbf{0.331} &   \textbf{0.392}   & \textbf{0.476}     \\
 \midrule
\multirow{2}{*}{3} & RSCD~\cite{lin2024robust}  &  0.548           & 0.240           &   0.365            & 0.385     \\
 & Ours                        &  \textbf{0.622}  & \textbf{0.280}  &   \textbf{0.396}   & \textbf{0.433}     \\
 \midrule
\multirow{2}{*}{5} & RSCD~\cite{lin2024robust}  &   0.468          & 0.203          &   0.360            & 0.344    \\
 & Ours                        &   \textbf{0.546} & \textbf{0.225}  &   \textbf{0.388}   & \textbf{0.386}    \\
 \midrule
\multirow{2}{*}{10} & RSCD~\cite{lin2024robust} &  0.395           & 0.224           &   0.354            & 0.324     \\
 & Ours                        &  \textbf{0.450}  & \textbf{0.245}  &   \textbf{0.388}   & \textbf{0.361}     \\
 
\bottomrule
\end{tabular}
\end{table}
\begin{table}[!t]
\centering
\setlength{\tabcolsep}{4pt}  
\caption{Experimental result under the SCD setup. 
Results obtained under the standard SCD setting of RSCD~\cite{lin2024robust}. 
Inference time is measured on the VL-CMU-CD dataset using a single NVIDIA RTX A5000 GPU.
$^{*}$ are taken from the RSCD paper and are scaled based on the FPS of RSCD for a fair comparison.
}
\label{tab:scd}
\begin{tabular}{cccccc}
\toprule
\multirow{2}{*}{Method} & \multicolumn{3}{c}{Dataset}      & \multirow{2}{*}{Average}    & \multirow{2}{*}{\shortstack{Inference\\time (ms)}}    \\
& VL-CMU-CD & PSCD  & ChangeSim \\             
\midrule
 DR-TANet~\cite{chen2021dr} & 0.607           &  0.023 &  -              &    -            & 31.84$^{*}$  \\
 C-3PO~\cite{wang2023reduce}    & \textbf{0.795}  &  0.048     &  -              &    -            & 23.54$^{*}$  \\
 RSCD~\cite{lin2024robust}     & \textbf{0.795}  &  0.337 &   0.344         &  0.492          & 31.14 \\ 
 Ours     & 0.791           &  \textbf{0.347}  &  \textbf{0.372}    &   \textbf{0.503} &   36.21 \\
 \bottomrule
\end{tabular}
\end{table}

\subsection{Experimental Results}
The experimental results are shown in \cref{tab:main}.
As shown in the table, our method consistently outperforms the state-of-the-art change detector, RSCD~\cite{lin2024robust}, across all datasets and database strides.
Additional implementation details are provided in the supplementary material.
In the most favorable case, database stride $s=1$, our method achieves an average performance of 0.476, significantly surpassing the RSCD~\cite{lin2024robust} of 0.429. 
As the reference database becomes sparser (\ie, as the database stride increases), the overall performance of both RSCD~\cite{lin2024robust} and our method degrades. However, the performance gap between the two remains consistent.
These results validate that our feature extraction and aggregation process enables effective change detection even under significant viewpoint misalignments. 

In addition, we also report the results under the conventional SCD setting in \cref{tab:scd}.
For evaluation, we follow the experimental protocol of RSCD~\cite{lin2024robust}, which represents the current state of the art in SCD.
To adapt our ECD pipeline to this setting, we set the database stride to 
$s=1$ and replace the reference subset with the ground-truth reference pair.
Even under this idealized condition, our method achieves the best average performance while remaining competitive with existing SCD approaches, demonstrating its strong capability for robust change detection.

\subsection{Ablation Study and Analysis Experiments}
We first conduct a component-wise ablation study to quantify the contribution of the main components of our method.
We then present a focused analysis of EMC from three complementary perspectives.
First, we compare EMC with alternative misalignment mitigation strategies based on single-reference warping. 
Second, we compare EMC with naive multi-view fusion strategies to examine whether the gain comes simply from using more references. 
Finally, we analyze the sensitivity of EMC to the quality of the first-stage VPR retrieval by replacing the retrieved references with oracle or distractor references.
Additional analyses of the remaining design components are provided in the supplementary material.

\begin{table}[!t]
\centering
\caption{Ablation study. We evaluate the impact of two key components in our method, the semantic aggregator and EMC. Results across three datasets show that combining both components yields the best results.
}
\label{tab:component}
\setlength{\tabcolsep}{5pt}
\begin{tabular}{ccccccc}
\toprule
\raisebox{-3pt}{\multirow{2}{*}{~~$s$~~}} & \multicolumn{2}{c}{Method} & \multicolumn{3}{c}{Dataset}  & \raisebox{-3pt}{\multirow{2}{*}{~Average~}}       \\
\cmidrule(lr){2-3} \cmidrule(lr){4-6}
                      & Agg. & EMC & VL-CMU-CD$^s$ & ~~PSCD$^s$~~ & ChangeSim$^s$ \\                  
\midrule
\multirow{4}{*}{1}
&              &              & 0.6107 & 0.2999 & 0.3688 & 0.4265 \\
& $\checkmark$ &              & 0.6112 & 0.3281 & 0.3621 & 0.4338 \\
&              & $\checkmark$ & 0.6990 & 0.2029 & \textbf{0.4036} & 0.4352 \\
& $\checkmark$ & $\checkmark$ & \textbf{0.7043} & \textbf{0.3308} & 0.3924 & \textbf{0.4758} \\
\midrule

\multirow{4}{*}{5}
&              &              & 0.4680 & 0.2027 & 0.3601 & 0.3436 \\
& $\checkmark$ &              & 0.4537 & 0.2023 & 0.3712 & 0.3424 \\
&              & $\checkmark$ & 0.5304 & 0.1285 & \textbf{0.3952} & 0.3514 \\
& $\checkmark$ & $\checkmark$ & \textbf{0.5456} & \textbf{0.2246} & 0.3879 & \textbf{0.3860} \\
\bottomrule
\end{tabular}
\end{table}

\begin{table*}[t]
\caption{Analysis experiments.  
\vspace{-10pt}
}
\label{tab:analyses}
\centering
\begin{subtable}[t]{0.32\linewidth}
\caption{Alternative misalignment mitigation modules.}
\label{tab:warp_method}
\setlength{\tabcolsep}{7pt}
\resizebox{1.00\linewidth}{!}{%
\begin{tabular}{@{}ccc@{}}
\toprule
VL-CMU-CD$^s$ &  $s$=1 & $s$=5    \\
\midrule
w/o align             &  0.619 &  0.468             \\
Homography            &  0.641 &  0.512             \\
Optical flow           &  0.604 & 0.467            \\
EMC (Ours)             &  \textbf{0.704} &  \textbf{0.546}            \\
\bottomrule
\end{tabular}
}
\end{subtable}\hfill
\begin{subtable}[t]{0.32\linewidth}
\centering
\setlength{\tabcolsep}{2pt}
\caption{Alternative multi-view fusion strategies.}
\label{tab:multiview_method}
\setlength{\tabcolsep}{7pt}
\resizebox{1.00\linewidth}{!}{%
\begin{tabular}{@{}ccc@{}}
\toprule
VL-CMU-CD$^s$ &  $s$=1 & $s$=5    \\
\midrule
Best single           &  0.619 & 0.468            \\
Average            &  0.586 &  0.382            \\
Concatenate           &  0.614 &  0.457            \\    
EMC (Ours) &  \textbf{0.704} & \textbf{0.546}             \\    
\bottomrule
\end{tabular}
}
\end{subtable} \hfill
\begin{subtable}[t]{0.32\linewidth}
\caption{Retrieval sensitivity analysis. \textit{O: Oracle, D: distractor}}
\label{tab:retrieval_sensitivity}
\setlength{\tabcolsep}{2.3pt}
\resizebox{1.00\linewidth}{!}{%
\begin{tabular}{@{}ccc@{}}
\toprule
VL-CMU-CD$^1$            &  F1 &  Patch acc.            \\
\midrule
1 O + 2 VPR                      &  \textbf{0.716} &  \textbf{97.71}            \\
3 VPR                                 &  0.704 &  97.57           \\
2 VPR    + 1 D               &  0.695 &  96.93            \\
1 VPR    + 2 D              &  0.663 &  95.58            \\  
\bottomrule
\end{tabular}
}
\centering
\end{subtable}
\end{table*}

\paragraph{Ablation study.} 
In \cref{tab:component}, we present a component-wise ablation study to evaluate the impact of each component of our proposed method.
When the environment mosaic composer is disabled, feature aggregation is performed using the features from the top-1 VPR result.
Conversely, when feature aggregation is disabled, the averaged mosaic environment representation $\tilde{f_r}$ is directly fed into the change detection module.
If both components are removed, features from the top-1 VPR result are used for change detection.
The results show that combining both aggregation and alignment yields the highest average scores, demonstrating their complementary nature.  

\paragraph{Analysis of misalignment mitigation.}
Our method employs a mosaic-based design to address misalignment. 
Beyond this choice, two straightforward alternatives can be considered: optical-flow-based warping and homography-based warping. 
To analyze the effectiveness of EMC, we construct two variants by replacing EMC with (i) RAFT-based optical-flow~\cite{teed2020raft} and (ii) a RANSAC-based homography~\cite{fischler1981random,sun2021loftr}, and report the results in \cref{tab:warp_method}. 
This comparison examines whether conventional single-reference geometric warping can provide benefits comparable to those of the proposed mosaic-based composition.
The results show that homography-based warping has only a marginal effect, whereas optical-flow-based warping degrades performance.
In contrast, EMC achieves the strongest performance among the compared misalignment mitigation strategies.

\paragraph{Analysis of multi-view fusion.}
We further compare EMC with alternative multi-view fusion strategies by replacing EMC with feature averaging and feature concatenation.
The results in \cref{tab:multiview_method} show that merely increasing the number of reference images does not necessarily improve performance. 
Naively mixing features can be harmful, especially at larger strides, where the retrieved subset is more likely to contain inaccurate references.
In contrast, EMC improves performance by spatially selecting reliable reference evidence rather than simply aggregating more reference images.

\paragraph{Analysis of retrieval sensitivity.}
Finally, we analyze the sensitivity of EMC to the quality of the first-stage VPR retrieval.
To this end, we replace the top VPR results with oracle and distractor references and report the results in \cref{tab:retrieval_sensitivity}. 
As expected, oracle references improve performance, while distractor references degrade performance by reducing valid pre-change evidence. 
For further analysis, we also report patch accuracy, which measures whether EMC selects patches from informative same-sequence references. 
Even under distractor injection, EMC maintains high patch accuracy, indicating that it largely avoids distractor patches and continues to exploit the remaining relevant references.

\subsection{Visualization of EMC}
\begin{figure*}[!t]
\centering
\includegraphics[width=1.0\linewidth]{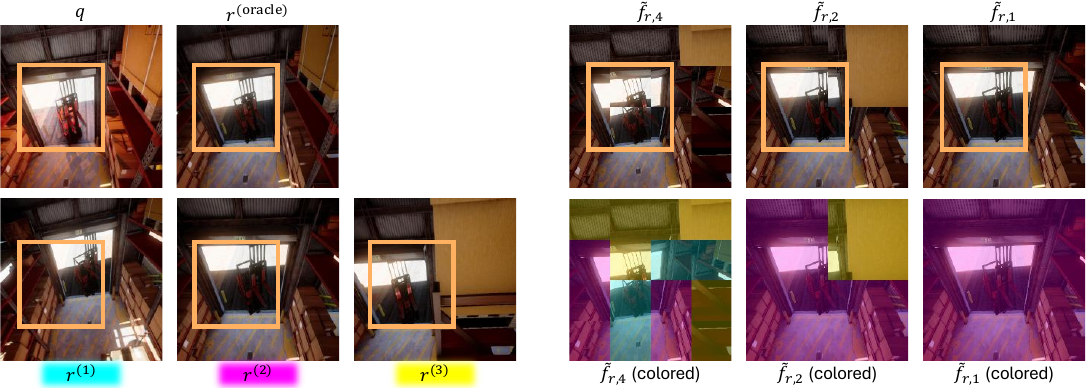}
\caption{
Visualizations of constructing a mosaic \( \tilde{f}_{r,n} \) from three reference images $r^{(1)}, r^{(2)}, r^{(3)}$. Appropriate patches are selected from each reference to form a mosaic-view similar to the ground-truth reference $r^{(\text{oracle})}$, particularly at the finer grid resolution ($n=4$).
For improved visibility, we also provide a colored version in which the color indicates the origin of each patch from the reference images.
}
\label{fig:main_mosaic_sample}
\end{figure*}

We present visual examples of the mosaic environment representations \( \{ \tilde{f}_{r,n} \} \) in \cref{fig:main_mosaic_sample}.
EMC operates on the features of the query $q$ and retrieved reference subset $ \{ r^{(1)}, r^{(2)}, r^{(3)} \}$, while $r^{(\text{oracle})}$ is not included in the reference database but is shown for illustrative purposes.
Although \( \{ \tilde{f}_{r,n} \} \) are feature representations, they are visualized as images by projecting the matched feature patches onto their spatial coordinates.
The visualization confirms that EMC can successfully imitate query views by selecting high-similarity patches from references for each query patch.
Additional examples and detailed analyses are provided in the supplementary material.

\subsection{Qualitative Results}
\begin{figure*}[!t]
\centering
\includegraphics[width=1.00\linewidth]{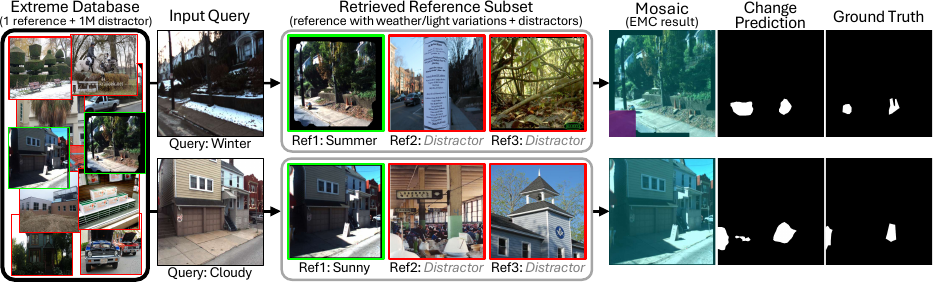}
\caption{
Qualitative results in a challenging setting with a large-scale reference database.
To simulate a less constrained real-world scenario, we restrict the reference database to one reference image per sequence and augment it with \(\mathcal{R}\)1M distractors.
For each query, the retrieved reference subset contains both true references and distractors, yet our method selectively extracts discriminative cues to achieve accurate predictions.
}
\label{fig:qualitative}
\end{figure*}

We present qualitative results in \cref{fig:qualitative} under a challenging evaluation setting based on VL-CMU-CD$^s$.
To construct a more practically challenging scenario beyond the standard benchmark, we retain only a single reference image from each test sequence (\(s=\inf\)) and further augment the resulting reference database with the $\mathcal{R}$1M distractor set~\cite{radenovic2018revisiting}. 
The \(\mathcal{R}\)1M distractor set comprises 1,001,001 distractor images for VPR, thereby enabling us to simulate ECD with a large-scale reference database containing diverse and unrelated images.

For the given query $q$, the retrieved reference subset includes both true reference and distractor images.
Leveraging the environment mosaic composer and semantic aggregator modules, our method selectively extracts relevant information from this subset, enabling accurate predictions.
Intermediate visualizations illustrating how these modules suppress incorrect VPR matches and compensate for camera viewpoint variations are provided in the supplementary material, along with additional qualitative examples on the standard benchmark datasets.

\section{Limitations}
Although ECD eliminates the need for predefined query-reference image pairs, it still requires sufficient field-of-view overlap between the query images and the images in the reference database to enable reliable change detection.
Additionally, this setting may increase the computational burden because relevant reference evidence must be discovered at inference time, whereas conventional SCD directly provides a predefined reference image for each query.
Representative failure cases are provided in the supplementary material.

\section{Conclusion}
In this paper, we introduce a new task termed Environmental Change Detection (ECD), which removes two unrealistic assumptions commonly made in conventional Scene Change Detection (SCD): the existence of predefined query-reference pairs and the availability of perfectly aligned reference images. 
ECD establishes a more challenging yet practical setting that better reflects real-world conditions.
We also propose the first solution to ECD.
Rather than processing the entire reference database, our method first constructs a compact reference subset by retrieving images relevant to the given query.
It then performs mosaic composition and semantic aggregation to reconstruct a reference scene corresponding to the query viewpoint.
The resulting representation preserves rich environmental information while remaining spatially aligned with the query, thereby providing a reliable basis for change detection.
Our method performs effectively in ECD and outperforms the evaluated SCD baselines.
Further experimental analyses demonstrate that our approach handles spatial misalignment more effectively than the alternative strategies.
We believe this work takes a significant step toward practical change detection under less constrained and more realistic conditions.

\section*{Acknowledgements}
This work was supported by Institute of Information \& communications Technology Planning \& Evaluation (IITP) grant funded by the Korea government(MSIT) (No.2022-0-01025, Development of core technology for mobile manipulator for 5G edge-based transportation and manipulation), and 
Korea Evaluation Institute Of Industrial Technology (KEIT) grant funded by the Korea government(MOTIE) (No.20023455, Development of Cooperate Mapping, Environment Recognition and Autonomous Driving Technology for Multi Mobile Robots Operating in Large-scale Indoor Workspace).

%
%
\bibliographystyle{splncs04}
\bibliography{main}

\clearpage
\newpage

\vbox{
  \vskip 0.1in
  \centering
  \LARGE \textbf{Supplementary Material}
  \vskip 0.1in
}

\appendix
\section{Implementation Details}
\subsection{ECD Setup Construction}
We describe the detailed procedure for preparing the experimental setup used in ECD.
Starting from the standard SCD datasets, which consist of triplets \((r, q, y)\), we first collect all reference images \(r\) and build the database. The query images \(q\) and their corresponding labels \(y\) are used without modification from the SCD setup.
To remove the assumption that a spatially aligned reference image \(r\) always exists for every query \(q\), we restrict the size of the database by applying a database stride \(s\). Specifically, only every \(s\)-th image in the sequence from the original database is included in the reference database \( \mathcal{I}_r \).  

Given that several SCD datasets (\eg, VL-CMU-CD~\cite{alcantarilla2018street} and ChangeSim~\cite{park2021changesim}) are composed of multiple sequences, we apply the stride on a per-sequence basis. Thus, only every \(s\)-th image within each sequence is included in the reference database \( \mathcal{I}_r \).
For PSCD~\cite{sakurada2020weakly}, where each panorama image is divided into 15 crops, we treat the set of cropped images from a single panorama as a sequence and apply the stride accordingly.

Formally, \( \mathcal{I}_r \) is defined as:
\begin{equation}
\mathcal{I}_r = \Big\{ r^{(m,j)} \,\Big|\, j \in \{1, 1+s, 1+2s, \dots\},\;  j  \leq L_m \;\text{ for each sequence } m \Big\} ,
\end{equation}
where \(L_m\) denotes the length of sequence \(m\).
This striding constraint reduces the spatial and temporal coverage of the reference database and removes the guarantee that a ground-truth reference-query pair exists.

\begin{figure*}[!t]
\centering
\includegraphics[width=0.98\linewidth]{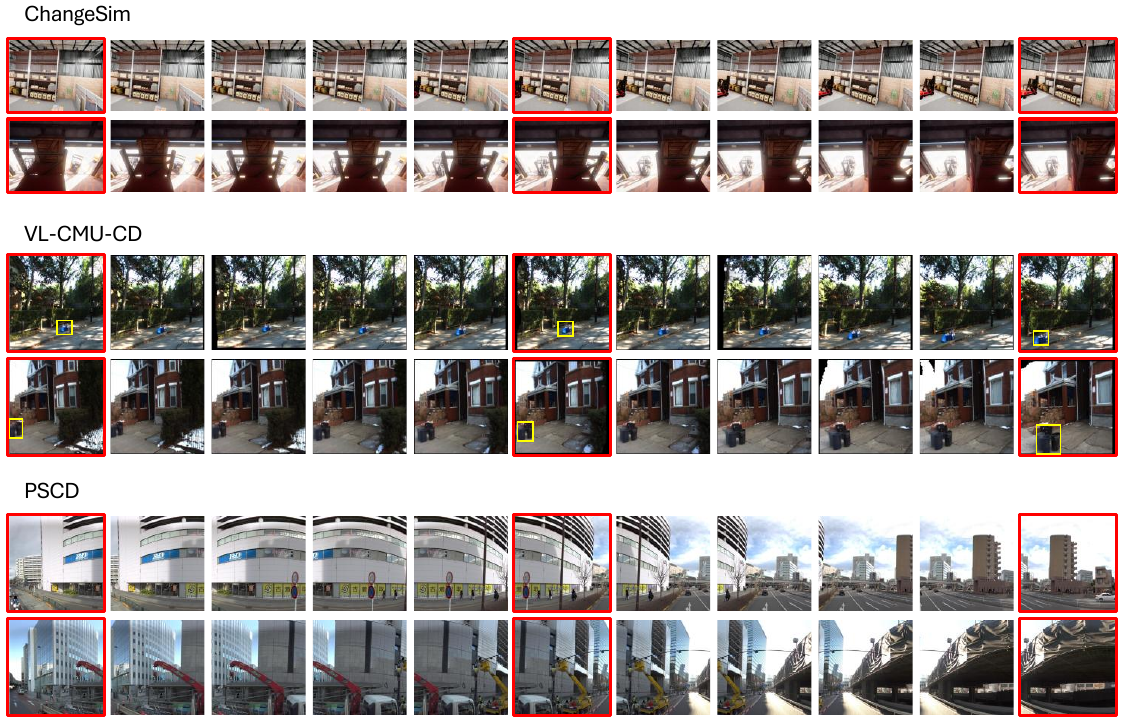}
\caption{Example images from the reference database. 
The figure compares the full reference images used in the conventional SCD setting with the strided reference database adopted in the proposed ECD setup.
In the SCD setting, all images are adopted as the reference, whereas in the ECD setup, only a subset of images is selected. 
The images outlined in red boxes indicate the reference images selected for the ECD reference database using a database stride of $s = 5$.
By applying a stride, ECD reduces the spatial and temporal density of the reference set, thereby intentionally introducing misalignment between queries and available references.  
Best viewed when zoomed in.
}
\label{fig:supp_taskexample}
\end{figure*}

The examples of the original and strided reference database are illustrated in~\cref{fig:supp_taskexample}.
As shown in the figure, increasing the stride sparsifies the reference set, reduces environmental coverage, and makes retrieval inherently more challenging.
This construction reflects real-world situations in which past environments are only sparsely recorded, such as camera-mounted vehicles that capture scenes intermittently.
The stride $s$ also acts as a controllable difficulty parameter, enabling consistent reproduction of experimental conditions and supporting systematic expansion of the benchmark (\eg, ChangeSim$^{1}$, ChangeSim$^{5}$, ChangeSim$^{10}$, \ldots).

\subsection{Architectural Details}
\paragraph{VPR module.}
We use a pre-trained BoQ model~\cite{ali2024boq} without any architectural modifications. 
The BoQ feature extractor is a pretrained ResNet-50~\cite{he2016deep}.

\paragraph{Semantic aggregator.}
We employ multi-head attention with 6 heads and a dropout rate of 0.1.
The feed-forward network (FFN) that follows the attention layer has a dimensionality of 384.

\paragraph{Change detection head.}
The change detection head operates on the reconstructed scene $f_r^*$, and the query feature $f_q$. The architecture of the change detection head is identical to that of RSCD~\cite{lin2024robust}.
First, two cross-attentions are applied in both directions from $f_r^*$ to extract information corresponding to $f_q$, and vice versa.
The resulting features are concatenated and passed through a sequence of layers: a $3 \times 3$ convolution that halves the channel dimension, followed by a $1 \times 1$ convolution to produce a single-channel change prediction map, and finally an upsampling layer. 
This process produces the final change detection output.

\subsection{Computational Resources}
The experiments were performed on a system equipped with the Intel Xeon Gold 6426Y CPU and a single NVIDIA RTX A5000 GPU. 
The software environment consisted of Python 3.10.16 and PyTorch 2.1.0, with CUDA version 12.2. 
The total GPU memory usage during the experiments remained below 24 GB.
Training was completed in under one day, whereas PSCD is reserved exclusively for testing and does not participate in training.

\section{VPR Retrieval Analysis}
We evaluate the top-1 retrieval accuracy of the VPR module to assess its effectiveness in retrieving relevant reference images under different database stride $s$ values.
To support this analysis, we define two levels of matching criteria: \textit{strict match} and \textit{coarse match}.
Note that these criteria are used exclusively for analysis in this section and are not incorporated into the main pipeline.

A \textit{strict match} is defined as follows: for VL-CMU-CD$^s$ and PSCD$^s$, it corresponds to the exact ground-truth reference pair; for ChangeSim$^s$, it refers to retrieved images within 1 meter and 10 degrees of the query, following SimSac~\cite{park2022dual}, which only considers distance.
A \textit{coarse match} is defined more loosely: for VL-CMU-CD$^s$ and PSCD$^s$, it includes any image from the same sequence as the ground-truth; for ChangeSim$^s$, it refers to retrieved images within 25 meters and 45 degrees of the query, following the retrieval metric~\cite{pion2020benchmarking}.

The results are presented in \cref{tab:vpr}. 
As the database stride $s$ increases, the retrieval accuracy for strict matches drops sharply, whereas coarse match rates remain relatively stable.  
This indicates that strictly relying on GT-aligned references becomes increasingly impractical at larger strides. Instead, models must leverage nearby images to support accurate predictions.
Notably, for VL-CMU-CD$^s$, the top-1 retrieval accuracy is lower compared to other datasets. This is because there is minimal viewpoint variation within each sequence, making many reference images visually similar and difficult to distinguish.

\begin{table*}[!t]
\centering
\setlength{\tabcolsep}{4pt}
\caption{The accuracy of the VPR module. Each value is the average accuracy (\%) computed on the training, validation, and test splits.}
\label{tab:vpr}
\begin{tabular}{ccccccccc}
\toprule
\raisebox{-3pt}{\multirow{2}{*}{\shortstack{~Database~\\stride $s$}}} & \multicolumn{2}{c}{ChangeSim$^s$} & \multicolumn{2}{c}{VL-CMU-CD$^s$} & \multicolumn{2}{c}{PSCD$^s$} & \raisebox{-3pt}{\multirow{2}{*}{\shortstack{Average\\(Strict)}}} & \raisebox{-3pt}{\multirow{2}{*}{\shortstack{Average\\(Coarse)}}} \\  
\cmidrule(lr){2-3} \cmidrule(lr){4-5} \cmidrule(lr){6-7}
&  ~Strict~ & Coarse & ~Strict~ & Coarse & ~Strict~ & Coarse & & \\
\midrule
 1  & 53.45 & 99.35 & 32.59 & 84.45 & 93.08 & 97.90 & 59.71 & 93.90 \\
 3  & 51.93 & 99.29 & 23.27 & 84.39 & 32.51 & 96.98 & 35.91 & 93.56 \\
 5  & 48.35 & 99.32 & 18.64 & 84.13 & 19.86 & 88.04 & 28.95 & 90.50 \\
 10 & 42.77 & 98.66 & 12.65 & 83.74 & 13.28 & 68.29 & 22.90 & 83.56 \\
\bottomrule
\end{tabular}
\end{table*}

\begin{figure*}[!t]
\centering
\includegraphics[width=1.00\linewidth]{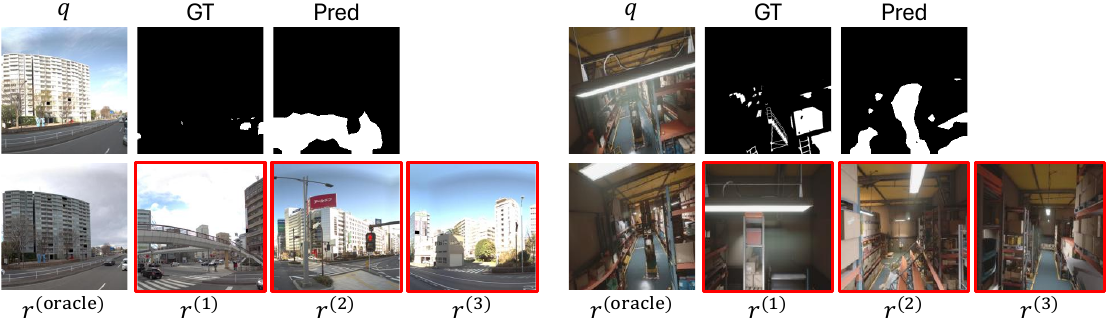}
\caption{Failure Analysis.
The first stage of our solution relies on constructing a reference subset using the VPR module. Consequently, when the VPR retrieval fails, the subsequent change detection process also fails due to the lack of relevant reference cues.
These observations reveal the limitations of the current pipeline and motivate future research on more robust or retrieval-independent reference database exploration strategies.
}
\label{fig:failure}
\end{figure*}

\section{Additional Analyses}

\paragraph{Reference subset size.} 
As shown in \cref{tab:nref}, the performance generally improves as the size of the reference subset $K$ (\ie, the number of reference images) increases, since a larger subset provides richer contextual cues.
However, the performance degrades when $K$ becomes too large, due to the inclusion of irrelevant or noisy references.  
This degradation becomes more pronounced as the database stride increases, because a larger stride yields a sparser reference database, making it easier for unrelated references to appear in $\mathcal{R}$ even with smaller $K$.  
Considering the trade-off between accuracy and computational efficiency, we set $K=3$ as the default configuration.

\paragraph{EMC grid resolution.}
In \cref{tab:grid}, we study how the choice of grid size in EMC affects performance.  
`Hier.' denotes a hierarchical configuration: hierarchical~$\times$2 averages representations obtained from 1$\times$1 and 2$\times$2 grids, while hierarchical~$\times$3 additionally incorporates a 4$\times$4 grid level.  
We observe that hierarchical configurations consistently outperform fixed single-scale grids, with hierarchical~$\times$3 achieving the highest average performance across both dataset stride settings.

\paragraph{Overlap-based adaptive striding.} 
To consider different viewpoint characteristics across datasets, we evaluate overlap-based adaptive striding. 
Specifically, we construct the reference database by selecting the first frame whose $r$-$r$ overlap falls below a predefined threshold.
\cref{tab:overlap_performance} reports the performance under this adaptive striding strategy. The table shows that the performance gaps across datasets are reduced.
This provides an alternative criterion for selecting a stride when balancing performance and database size.

\paragraph{PSCD fine-tuning protocol.} 
Although RSCD~\cite{lin2024robust} primarily treats PSCD~\cite{sakurada2020weakly} as a test-only benchmark, it additionally reports results under a PSCD fine-tuning protocol. For a more comprehensive evaluation, we adopt the same protocol and present the corresponding results in \cref{tab:PSCD_finetune}. Our method also achieves higher performance than RSCD under this setting.

\begin{table}[!t]
\centering
\caption{Analysis experiment on the size of the reference subset $K$. 
Performance improves as $K$ increases, but degrades when excessive references are included.}
\label{tab:nref}
\setlength{\tabcolsep}{7pt}
\begin{tabular}{cccccc}
\toprule         
$s$ & $K$ & VL-CMU-CD$^s$ & ~~PSCD$^s$~~ & ChangeSim$^s$ & Average         \\
\midrule
\multirow{4}{*}{1}
&1             &  0.6282            &  0.3300            &  0.3889            &  0.4490          \\
&3             &  0.7043            &  0.3308            &  0.3924            &  0.4758          \\
&5             &  0.7100            &  \textbf{0.3397}   &  0.3940            &  0.4812          \\
&10            &  \textbf{0.7281}  &  0.3255            &  \textbf{0.3950}   &  \textbf{0.4829} \\
\midrule
\multirow{4}{*}{5}
&1             &  0.4908            &  \textbf{0.2678}   &  0.3855            &  0.3814          \\
&3             &  \textbf{0.5456}   &  0.2246            &  0.3879            &  \textbf{0.3860}   \\
&5             &  0.5203            &  0.2325            &  \textbf{0.4010}   &  0.3846       \\
&10            &  0.5235            &  0.2316            &  0.3984            &  0.3845          \\
\bottomrule
\end{tabular}
\end{table}

\begin{table}[!t]
\centering
\caption{ 
Analysis experiment on the grid size of the environment mosaic composer. Hierarchical settings generally show improved performance, with ‘Hier. $\times$3’ achieving the best result.
}
\label{tab:grid}
\setlength{\tabcolsep}{7pt}
\begin{tabular}{cccccc}
\toprule
$s$ & Grid size & VL-CMU-CD$^s$ & ~~PSCD$^s$~~ & ChangeSim$^s$ & Average        \\
\midrule
\multirow{5}{*}{1} 
& 1$\times$1             & 0.6929           &   0.3339           &   0.3794           &   0.4687           \\        
& 2$\times$2             & 0.6884           &   0.2651           &   0.3849           &   0.4461           \\        
& 4$\times$4             & 0.5645           &   0.2623           &   \textbf{0.3930}  &   0.4066           \\        
& Hier. $\times$2        & 0.6957           &   \textbf{0.3446} &   0.3826           &   0.4743           \\        
& Hier. $\times$3        & \textbf{0.7043}  &   0.3308           &   0.3924           &   \textbf{0.4758}  \\
\midrule
\multirow{5}{*}{5} 
& 1$\times$1             & 0.4755           &   \textbf{0.2562}  &   0.3790           &   0.3702            \\        
& 2$\times$2             & 0.4986           &   0.1790           &   0.3863           &   0.3546            \\        
& 4$\times$4             & 0.4089           &   0.1197           &   0.3851           &   0.3046            \\        
& Hier. $\times$2        & 0.5181           &   0.2230           &   0.3868           &   0.3760            \\        
& Hier. $\times$3        & \textbf{0.5456}  &   0.2246           &   \textbf{0.3879}  &   \textbf{0.3860}   \\        
\bottomrule
\end{tabular}
\end{table}

\begin{table*}[t]
\caption{Analysis experiments.  \vspace{-10pt}}
\label{tab:analyses}
\centering
\begin{subtable}[t]{0.59\linewidth}
\centering
\setlength{\tabcolsep}{10pt}
\caption{Performance under adaptive striding based on $r$-$r$ overlap thresholds.}
\label{tab:overlap_performance}
\setlength{\tabcolsep}{3pt}
\resizebox{1.00\linewidth}{!}{%
\begin{tabular}{c|ccc}
\toprule
overlap &  VL-CMU-CD & ~~~PSCD~~~ &  ChangeSim       \\
\midrule
95.0\% & 0.4420 & - & 0.3709 \\
85.0\% & 0.3696 & - & 0.3646 \\
75.0\% & 0.3543 & 0.2684 & 0.3581 \\
\bottomrule
\end{tabular}
}
\end{subtable} \hfill
\begin{subtable}[t]{0.39\linewidth}
\caption{SCD results under the PSCD fine-tuning setting.}
\label{tab:PSCD_finetune}
\setlength{\tabcolsep}{10pt}
\resizebox{1.0\linewidth}{!}{%
\begin{tabular}{cc}
\toprule
Method &PSCD$^{\rm{finetune}}$  \\             
\midrule
 DR-TANet~\cite{chen2021dr}          &  0.190  \\
 C-3PO~\cite{wang2023reduce}     &  0.433      \\
 RSCD~\cite{lin2024robust}       &  0.442 \\ 
 Ours              &  \textbf{0.446}  \\
 \bottomrule
\end{tabular}
}
\centering
\end{subtable}
\end{table*}

\section{Failure Analysis}
To further complement our evaluation, we present representative failure cases in \cref{fig:failure}.
These examples show typical errors, mainly caused by inaccurate retrieval in the VPR stage.
The first stage of our solution relies on constructing a reference subset using the VPR module. 
When VPR retrieval fails, the subsequent change detection process also fails.
These failure cases motivate future research on alternative ways of exploring the environment database.

\section{Visualization and Analysis of EMC}
\begin{figure*}[!t]
\centering
\includegraphics[width=1.0\linewidth]{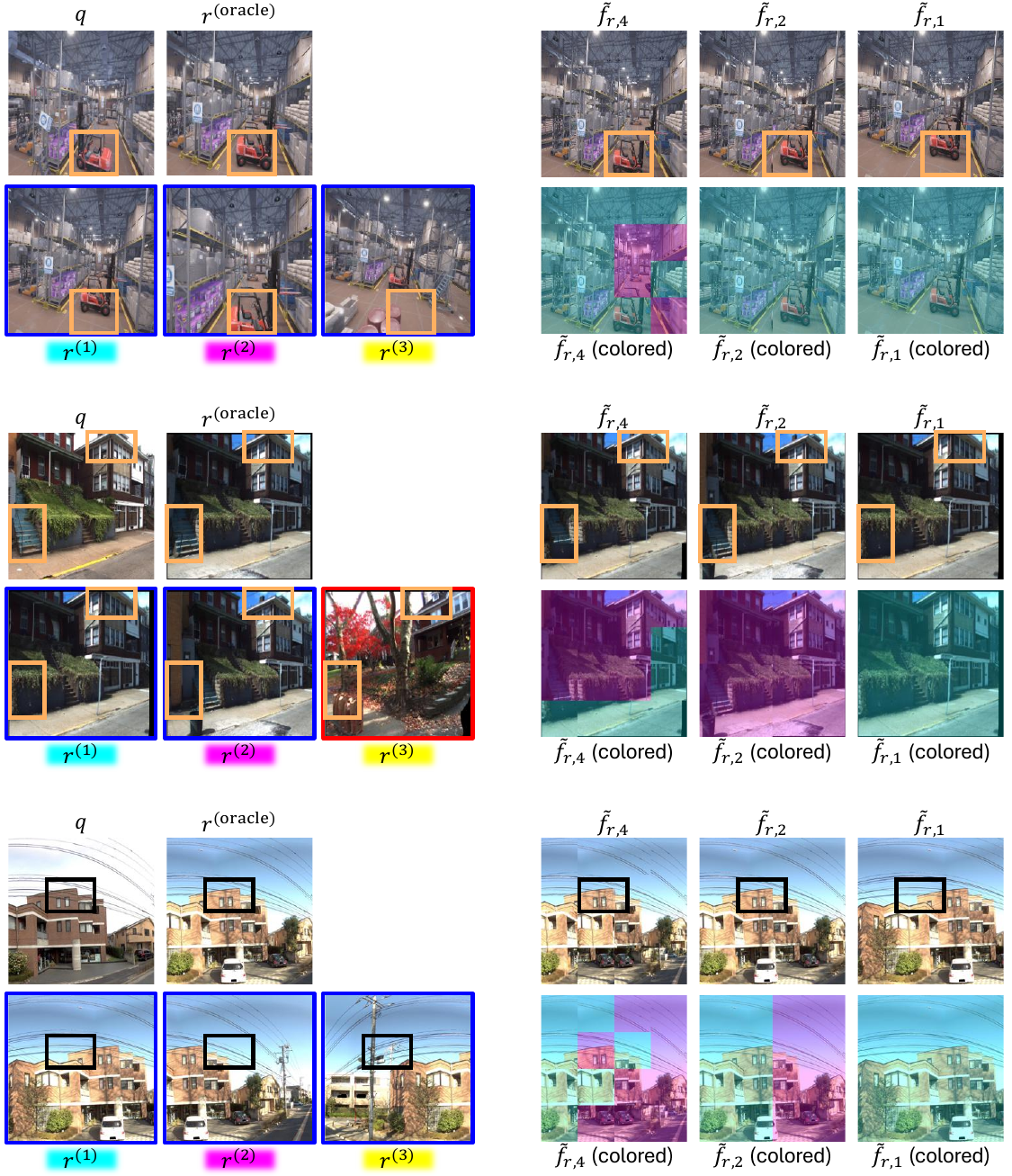}
\caption{Examples of mosaic environment representations. 
Given a query image $q$, EMC generates mosaic environment representations at multiple grid levels using reference images $r^{(1)}$, $r^{(2)}$, and $r^{(3)}$.  
$r^{(\text{oracle})}$ is shown for reference purposes.  
At finer grid levels, the generated mosaics closely imitate the viewpoint of the query.  
The borders of the reference images are colored to indicate match quality: green for strict matches, blue for coarse matches, and red for incorrect matches.
}
\label{fig:supp_align_a}
\end{figure*}

\begin{figure*}[!t]
\centering
\includegraphics[width=1.0\linewidth]{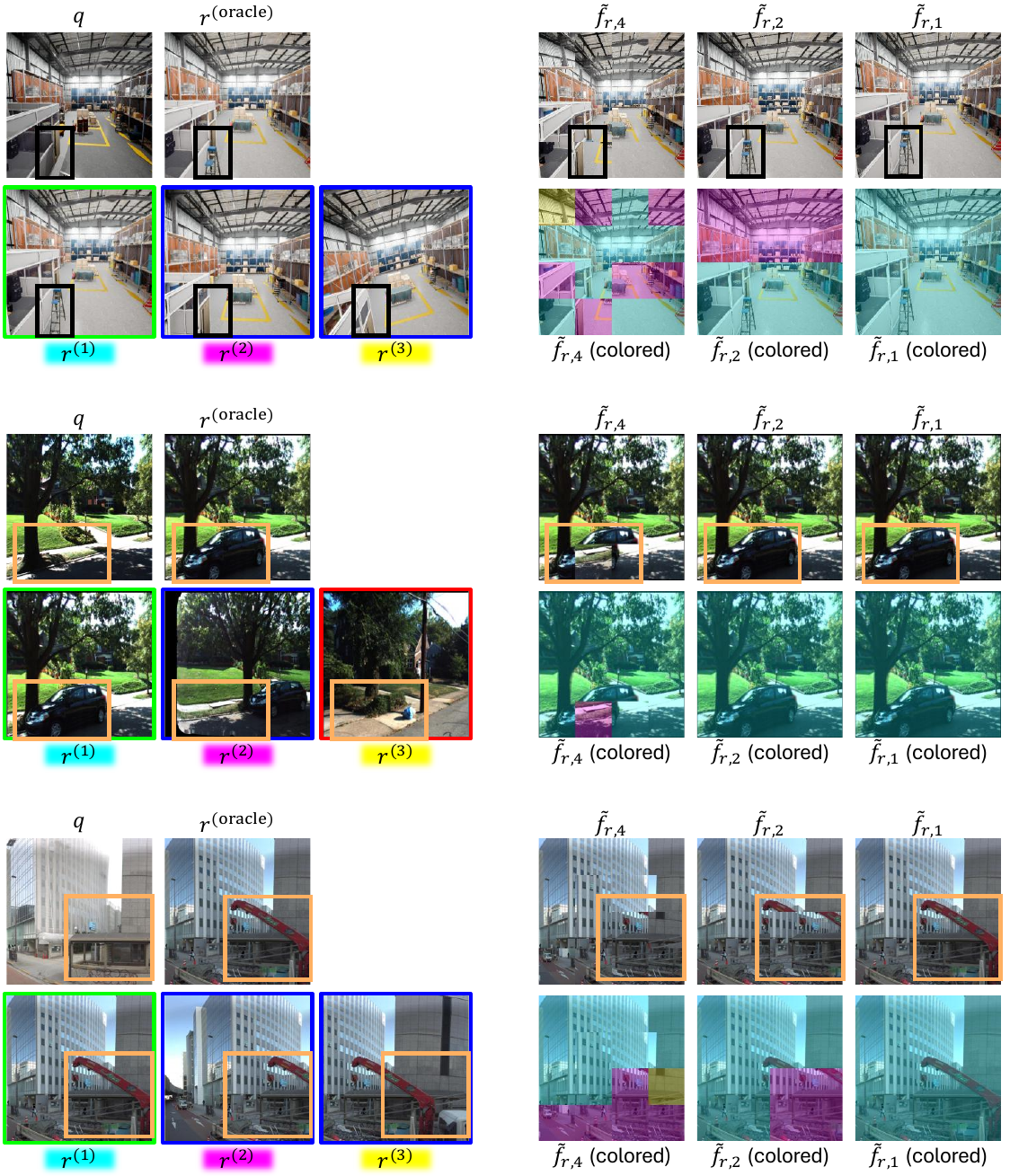}
\caption{Examples of mosaic environment representations under large object changes. 
When a substantial change spans the entire area of a patch, the mosaic environment representation at fine grid levels may fail to align accurately. 
However, the mosaic environment representation at coarse grid levels remains robust and provides strong evidence for the effectiveness of the hierarchical approach.
}
\label{fig:supp_align_b}
\end{figure*}

\begin{figure*}[!t]
\centering
\includegraphics[width=1.0\linewidth]{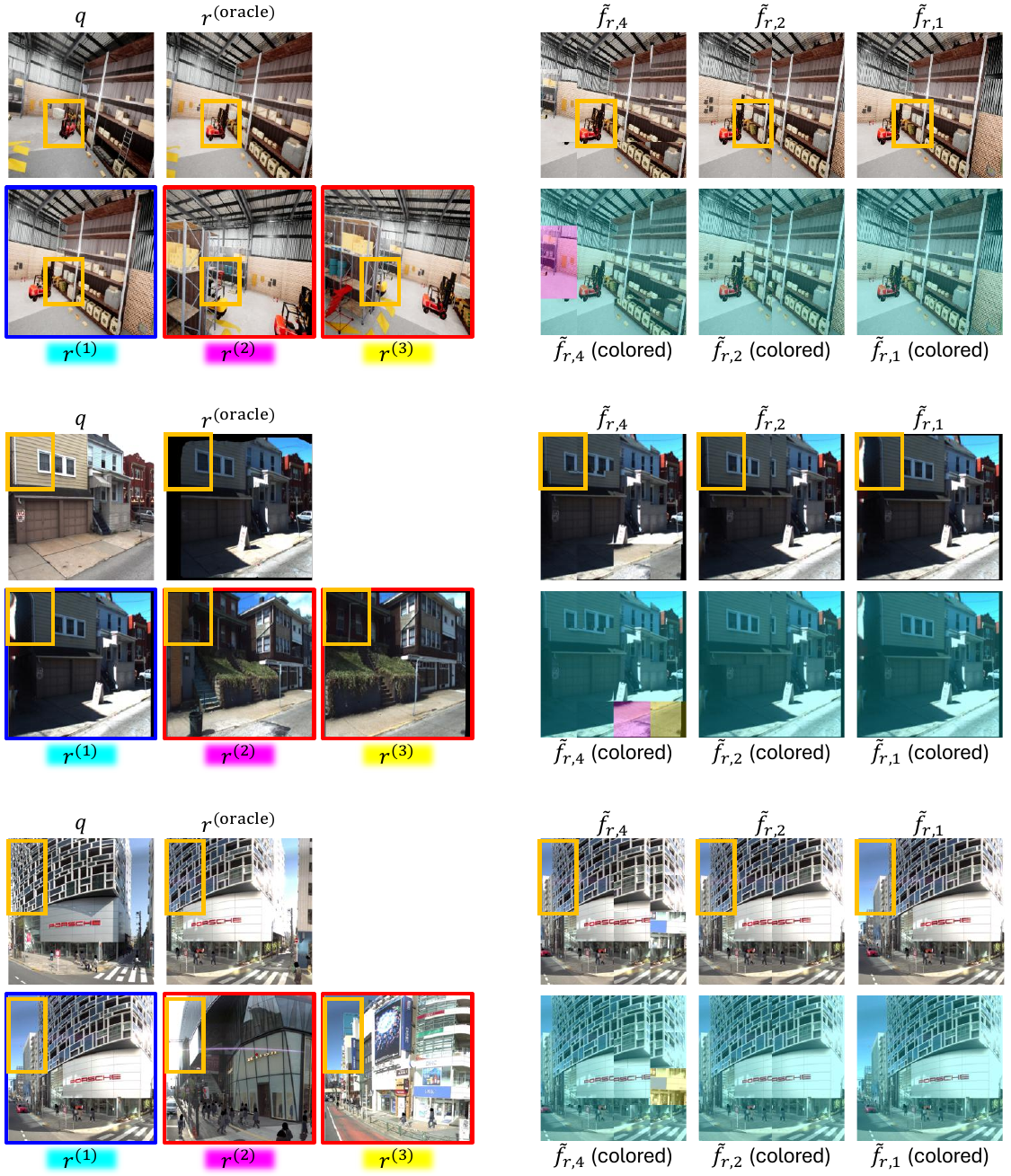}
\caption{Examples of mosaic environment representations under a large database stride and incorrect retrieval.  
Even when the reference subset $\mathcal{R}$ contains multiple incorrect references, EMC effectively suppresses the influence of unrelated images and reconstructs the mosaic environment representation primarily using the most relevant match.
}
\label{fig:supp_align_c}
\end{figure*}

We provide visual examples in \cref{fig:supp_align_a,fig:supp_align_b,fig:supp_align_c} to showcase the effectiveness of our proposed EMC in producing hierarchical mosaic environment representations.
In the visual examples, $q$ denotes the query image, $r^{(\text{oracle})}$ is the ground truth reference image, and $r^{(k)} \in \mathcal{R}$ are the retrieved reference subset $\mathcal{R}$. 
Note that $r^{(\text{oracle})}$ is not guaranteed to be included in the reference database depending on the query and database stride, but is shown here for illustrative purposes.
The retrieved reference images are visually marked as follows: green borders indicate a strict match, blue borders indicate a coarse match, and red borders indicate incorrect matches.

Although \( \{ \tilde{f}_{r,n} \} \) are feature representations, we visualize them as images by projecting the matched feature patches onto their spatial coordinates.
For improved visibility, we also provide a colored version in which the color indicates the origin of each patch from the reference images. 
Specifically, $r^{(1)}$, $r^{(2)}$, and $r^{(3)}$ are colored in cyan, magenta, and yellow, respectively.
To aid in understanding viewpoint alignment, corresponding regions are highlighted with black or orange boxes.

\cref{fig:supp_align_a} presents examples of mosaic environment representations.  
These results show that our method can imitate the query view by selecting high-similarity patches for each query patch from multiple references.
In particular, the mosaic environment representations on the $4 \times 4$ grid effectively approximate the target viewpoints.

\cref{fig:supp_align_b} illustrates a case involving changes to a large object.
Because EMC operates at the patch-level similarity, alignment accuracy may degrade at finer grid resolutions, especially when a patch is entirely altered.
Even so, the hierarchical design maintains robust alignment at coarser grid levels.
Thus, large changes and substantial scene variations are still handled effectively.
These results highlight the complementary roles of fine- and coarse-level alignment, validating our hierarchical design.

\cref{fig:supp_align_c} depicts a challenging case with a large database stride $s$, where no correctly aligned reference images are available and many incorrect distractors are included in the reference subset.
Even under these conditions, EMC effectively suppresses the influence of unrelated images and reconstructs the view primarily using the relevant match.

Overall, these results provide strong evidence that EMC can operate reliably across a wide range of conditions.
However, the outputs may show semantic inconsistencies, particularly near patch boundaries.
Therefore, we leverage the strong spatial alignment provided by the mosaic representation to employ the following semantic aggregator to integrate the reference features.

\section{Qualitative Results}
\begin{figure*}[!t]
\centering
\includegraphics[width=1.0\linewidth]{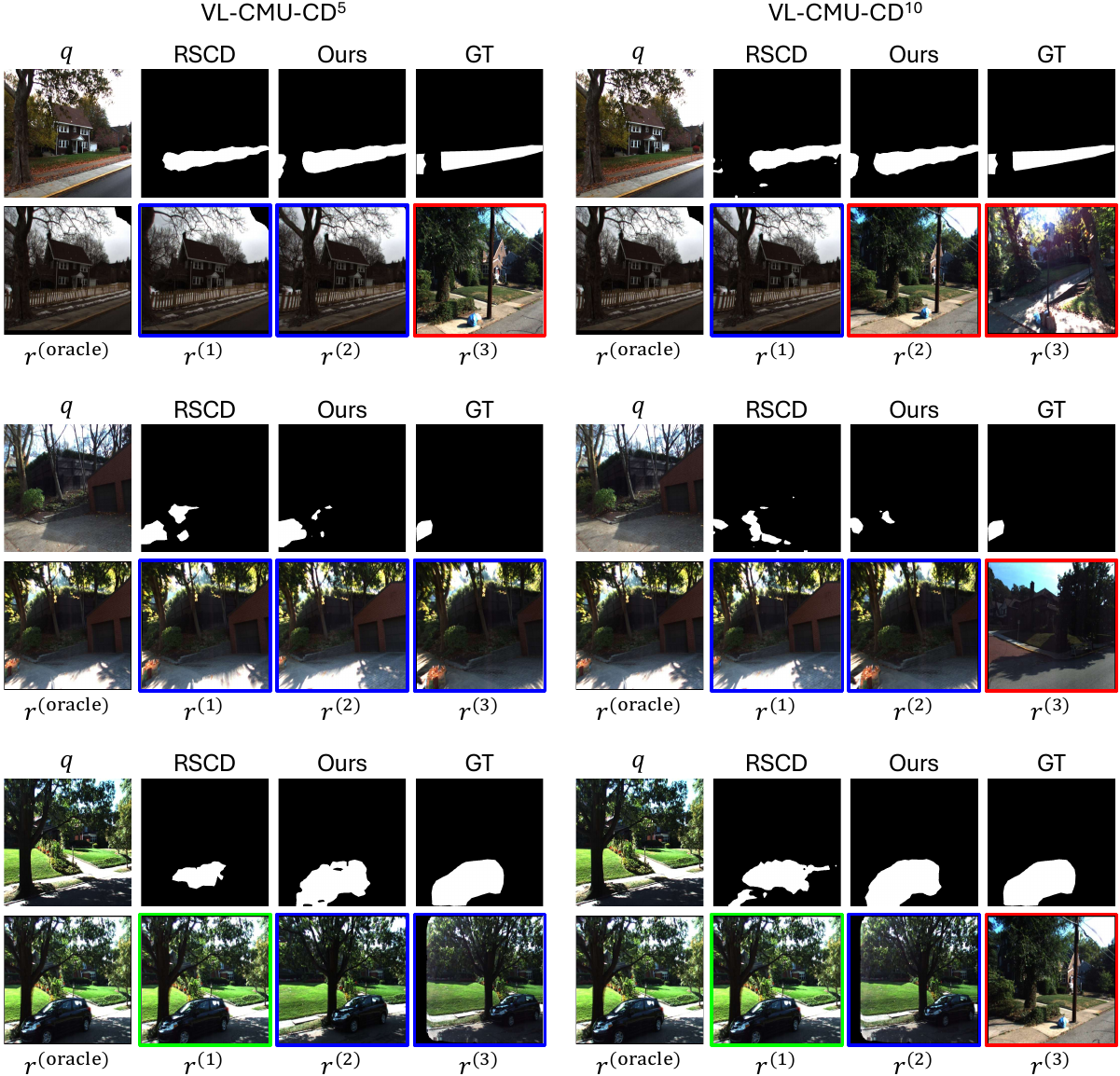}
\caption{Qualitative results on VL-CMU-CD$^5$ and VL-CMU-CD$^{10}$. 
}
\label{fig:supp_result_vlcmu}
\end{figure*}

\begin{figure*}[!t]
\centering
\includegraphics[width=1.0\linewidth]{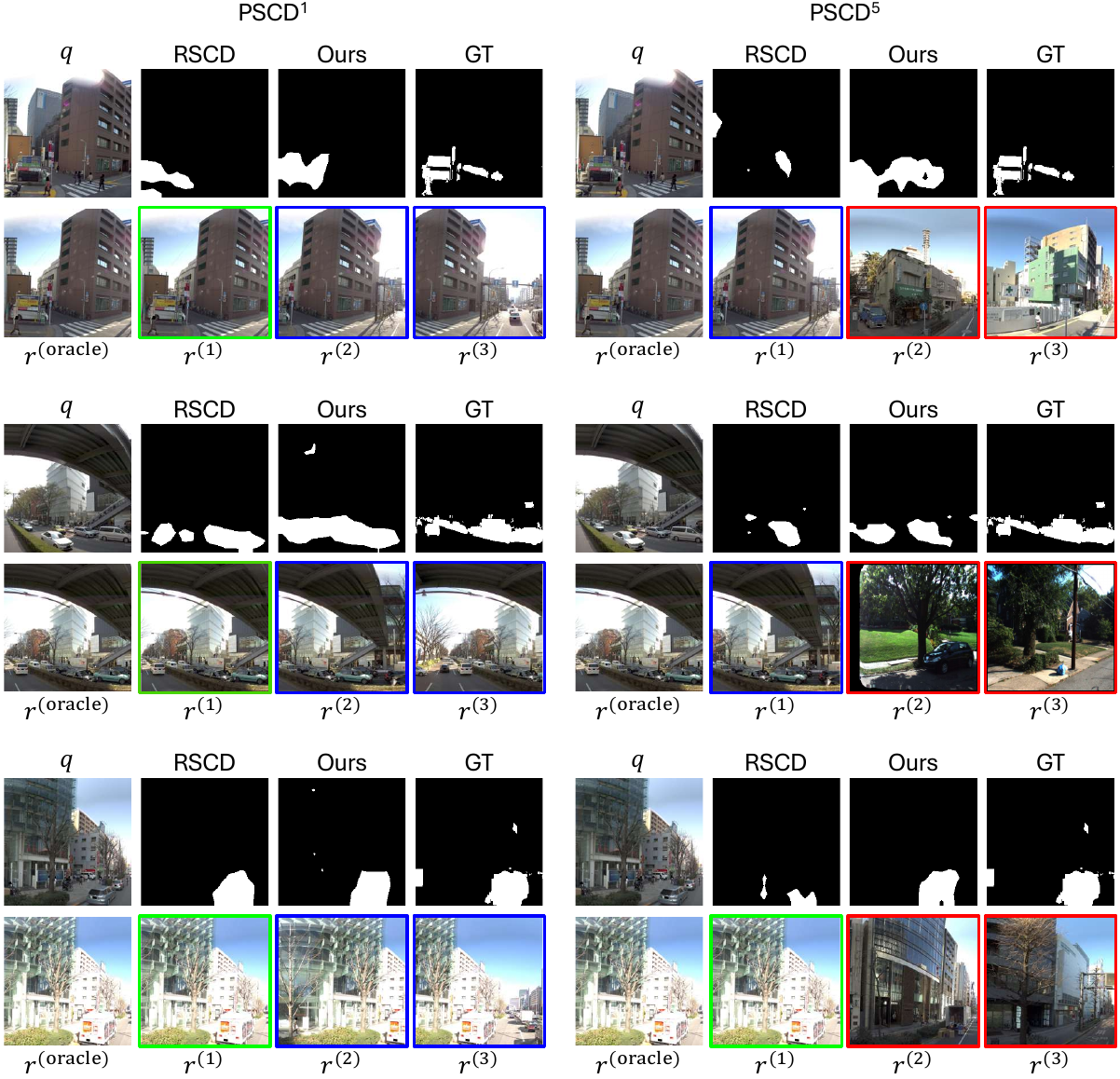}
\caption{Qualitative results on PSCD$^1$ and PSCD$^5$. 
}
\label{fig:supp_result_pscd}
\end{figure*}

\begin{figure*}[!t]
\centering
\includegraphics[width=1.0\linewidth]{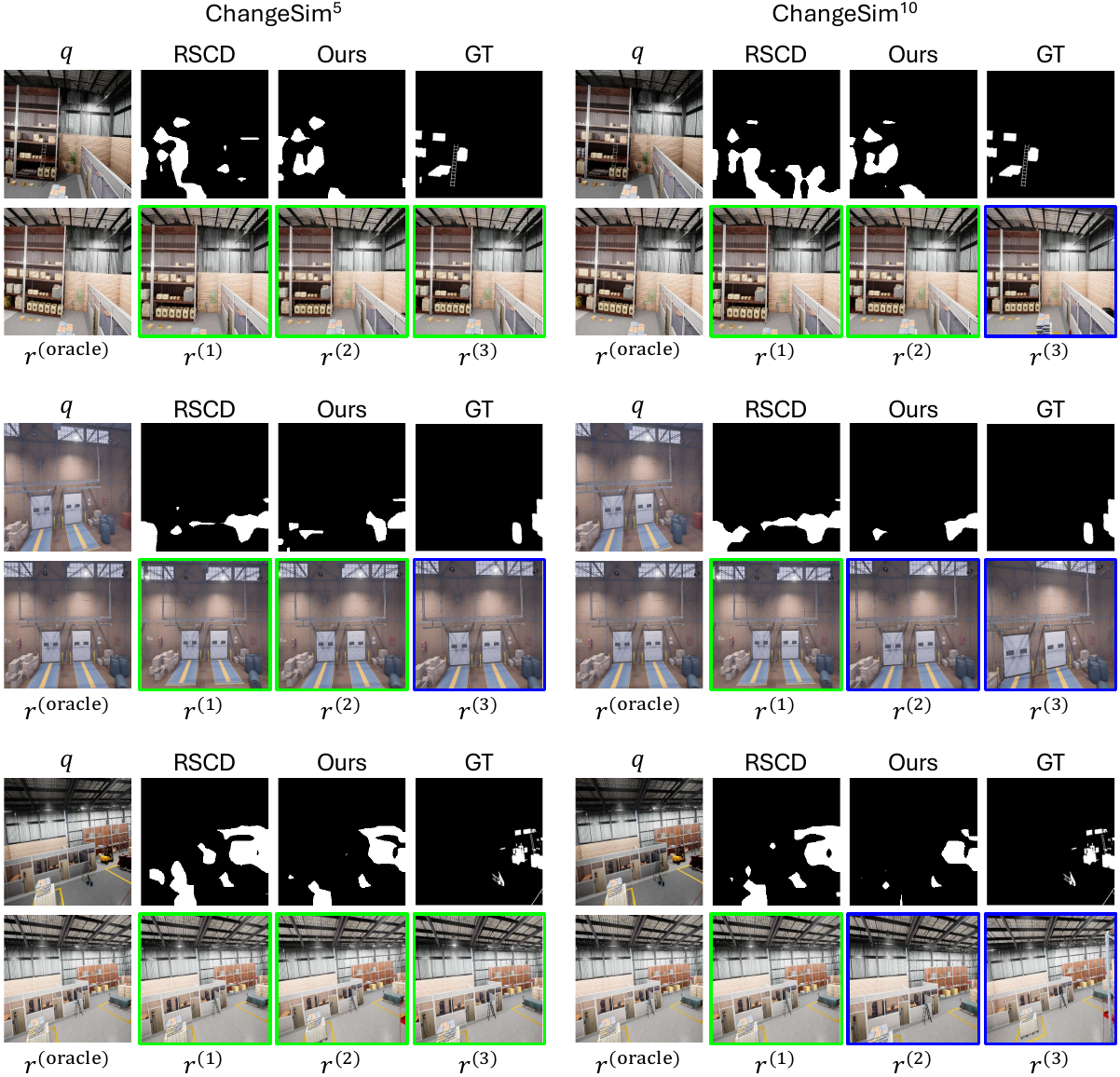}
\caption{Qualitative results on ChangeSim$^5$ and ChangeSim$^{10}$. 
}
\label{fig:supp_result_changesim}
\end{figure*}

We present additional qualitative results across various datasets in \cref{fig:supp_result_changesim,fig:supp_result_vlcmu,fig:supp_result_pscd}.
We provide prediction results for each query image $q$ under two different dataset stride $s$ settings, to illustrate how spatial information and prediction difficulty vary depending on the sparsity of the reference database.  

As in the previous section, $r^{(\text{oracle})}$ is the ground truth reference image, and $r^{(k)} \in \mathcal{R}$ are the retrieved reference subset $\mathcal{R}$.
The borders of the reference images are colored to indicate match quality: green for strict matches, blue for coarse matches, and red for incorrect matches.

Our method achieves accurate predictions even without exact viewpoint matches by leveraging nearby frames. 
Moreover, it demonstrates greater robustness to viewpoint changes compared to the baseline. 
Even when the VPR retrieval includes incorrect matches, most notably observed in the PSCD$^5$, our model can still produce accurate predictions, thanks to effective attention mechanisms.

\section{Broader Impacts}
SCD and the proposed ECD can be applied to a wide range of real-world scenarios, including natural disaster assessment, urban development monitoring, and automated warehouse management. These technologies have the potential to enhance safety, efficiency, and environmental awareness.
However, their deployment may raise privacy concerns and enable intrusive surveillance applications. Care must be taken to ensure that such systems are developed and used in a way that respects individual rights and complies with legal and ethical standards.

\clearpage

\end{document}